\pdfoutput=1

\documentclass[11pt]{article}

\usepackage[final]{acl}

\usepackage{times}
\usepackage{latexsym}

\usepackage[T1]{fontenc}

\usepackage[utf8]{inputenc}

\usepackage{microtype}

\usepackage{inconsolata}

\usepackage{graphicx}

\usepackage{subcaption}
\usepackage{color}
\definecolor{dkgreen}{rgb}{0,0.6,0}
\definecolor{gray}{rgb}{0.5,0.5,0.5}
\definecolor{mauve}{rgb}{0.58,0,0.82}
\usepackage{listings}
\usepackage{amsmath}
\usepackage[colorinlistoftodos]{todonotes}

\usepackage{paralist}
\usepackage{url}
\usepackage{hyperref}
\usepackage{enumitem}

\title{Fair Play in the Newsroom:\\Actor-Based Filtering Gender Discrimination in Text Corpora}

\author{
 \textbf{Stefanie Urchs\textsuperscript{1,2}},
 \textbf{Veronika Thurner\textsuperscript{1}},
 \textbf{Matthias Aßenmacher\textsuperscript{2,3}},
 \textbf{Christian Heumann\textsuperscript{2}},
\\
 \textbf{Stephanie Thiemichen\textsuperscript{1}}
 \\
\\
 \textsuperscript{1}Faculty for Computer Science and Mathematics,\\ Hochschule München University of Applied Sciences,
 \textsuperscript{2}Department of Statistics, LMU Munich,\\
 \textsuperscript{3}Munich Center for Machine Learning (MCML), LMU Munich
\\
 \small{
   \textbf{Correspondence:} \href{mailto:stefanie.urchs@hm.edu}{stefanie.urchs@hm.edu}
 }
}

\begin{document}
\maketitle

\begin{abstract}
Language corpora are the foundation of most natural language processing research, yet they often reproduce structural inequalities. One such inequality is gender discrimination in how actors are represented, which can distort analyses and perpetuate discriminatory outcomes. This paper introduces a user-centric, actor-level pipeline for detecting and mitigating gender discrimination in large-scale text corpora. By combining discourse-aware analysis with metrics for sentiment, syntactic agency, and quotation styles, our method enables both fine-grained auditing and exclusion-based balancing. Applied to the \texttt{taz2024full} corpus of German newspaper articles (1980–2024), the pipeline yields a more gender-balanced dataset while preserving core dynamics of the source material. Our findings show that structural asymmetries can be reduced through systematic filtering, though subtler biases in sentiment and framing remain. We release the tools and reports to support further research in discourse-based fairness auditing and equitable corpus construction.
\end{abstract}

\section{Introduction}
Large-scale text corpora are central to natural language processing and related fields, yet they often reproduce societal inequalities. Wikipedia reflects gender imbalances in coverage \cite{Wagner2021-sk}, job advertisements use gendered wording that reinforces hierarchies \cite{gaucher2011evidence}, and the film industry promotes stereotypes \cite{Kagan2020-tq}. Such examples show how corpora encode and normalise discrimination in persistent ways. Detecting these patterns is essential, but given the scale of modern datasets, manual inspection is infeasible. Automatic methods are needed to reveal structural inequalities at scale, and crucially, detection must be paired with curation: once problematic material is identified, corpora should be rebalanced to provide more reliable input data for NLP applications and more trustworthy resources for research.

\citet{urchs2025taz2024fullanalysinggermannewspapers} introduced a linguistically grounded pipeline to detect gender discrimination in German newspapers through actor-level discourse analysis, examining how named actors are represented via nomination and predication. Building on this work, we extend the pipeline to enable both fine-grained fairness auditing and corpus-level discrimination reduction. Our contributions are:

\begin{compactenum}
\item Novel actor-level discrimination markers, including syntactic roles, quote attribution, and sentiment bias.
\item Structured, human-readable reports that support qualitative and diachronic analysis.
\item A method for generating gender-balanced corpora by excluding disproportionately discriminatory texts.
\item An open-source release of the pipeline to ensure transparency, reproducibility, and collaboration.
\end{compactenum}

This paper offers tools and insights for creating fairer corpora by revealing how social groups are represented in text. We combine discourse-informed analysis with scalable processing to enable actor-level discrimination detection and targeted corpus balancing.

\section{Related Work and Conceptual Background}
Detecting gender discrimination in text requires an interdisciplinary foundation that integrates perspectives from linguistics, gender studies, and computer science. 

\subsection{Gender and Linguistic Discrimination}

In this work, we adopt a differentiated understanding of gender and discrimination that draws from linguistic discourse analysis, gender studies, and computational fairness research.

\textbf{Gender} is treated here as a socially constructed identity rather than a fixed biological or grammatical category. While linguistic gender follows grammatical rules~\cite{Kramer_2020}, and NLP research often reduces gender to binary labels~\cite{devinney2022theories}, we work with the broader notion of \textit{social gender}, which is fluid, contextual, and shaped through interaction~\cite{west1987doing}. Our empirical analysis is restricted to binary categories because the corpus lacks sufficient non-binary representation, but the approach can be adapted to encompass more inclusive forms of gender representation.

\textbf{Discrimination}, in contrast to bias or fairness, is understood here as a social effect: the observable outcome of differential treatment based on protected attributes such as gender. Following \citet{reisigl2017}, we view social discrimination as a process that disadvantages individuals through recurring patterns in language. This perspective differs from many machine learning approaches, where \textit{bias} is framed as statistical imbalance and \textit{fairness} as compliance with formal metrics such as demographic parity or equal opportunity \cite{blodgett2020language,caton_fairness_2024}. While effective for measuring distributional disparities, these frameworks largely ignore semantic and discursive aspects of language, where subtle forms of discrimination are often embedded.

\subsection{Computational Discrimination Detection}

In computational research, discrimination is usually formalised through fairness metrics such as demographic parity, equalised odds, or individual fairness~\cite{10.1145/3457607}. While these approaches are scalable and reproducible, they treat social categories as fixed attributes and largely abstract away from semantics and discourse~\cite{blodgett2020language}. Applied to text, this has produced methods for hate speech detection, sentiment disparity, or stereotyping, typically relying on keyword lists or supervised classifiers. Such methods yield valuable insights but operate mainly at the document level, labelling texts as``discriminatory'' or ``non-discriminatory'' and overlooking how unequal treatment is distributed within discourse.

In contrast, \citet{urchs2025taz2024fullanalysinggermannewspapers} proposed an actor-level approach that identifies individuals and analyses how they are represented through \textit{nomination} and \textit{predication}. By shifting the focus from entire texts to the representation of actors within them, this perspective reveals structural asymmetries that remain invisible to classical bias-detection methods.

\subsection{Actor-Level Discrimination Detection Pipeline}
\label{sec:pipeline}
Our pipeline builds on prior work by \citet{urchs-etal-2024-detecting,urchs2025taz2024fullanalysinggermannewspapers}. The first paper introduces actor-based fairness analysis in isolated English texts using a modular pipeline that combines information extraction with discourse analysis. It detects gender discrimination at the actor level by identifying \textit{nomination} and \textit{predication}, extracting actors via named entity recognition (NER), resolving pronouns through coreference, and storing references (names, titles, generic forms) in a structured knowledge base. For each actor (per text), all sentences in which they are mentioned are analysed for sentiment, gender-coded language, and framing. The resulting discrimination report provides per-text metrics such as:

\begin{compactitem}
    \item \textbf{Actor counts}: Number of distinct male-, female-, non-binary- and undefined-coded actors per text.
    \item \textbf{Mention counts}: Total number of pronoun or name-based references per gender group.\footnote{The difference between actor counts and mention counts can be illustrated with a simple example: a text with one male actor mentioned ten times differs from a text with ten female actors each mentioned once. Both cases result in ten actor references, but the distribution of visibility is fundamentally different.}
    \item \textbf{Sentiment}: Average sentiment score of all predications linked to each actor or gender group.
    \item \textbf{Gender-coded language}: Count of feminine-coded and masculine-coded terms in predications, based on lexicons from \citet{gaucher2011evidence}.
\end{compactitem}

The second paper scales this analysis to the \texttt{taz2024full} corpus (1.8M German newspaper articles, 1980–2024). It adapts the pipeline for German, replaces the sentiment model with a BERT-based classifier trained on German news, and adds markers for gender-neutral language and generic masculine usage. Actor-level metrics are aggregated by year, enabling longitudinal analysis of representation and framing. Additional features include:

Beyond the metrics introduced in the earlier paper, the \texttt{taz2024full} version adds:
\begin{compactitem}
    \item \textbf{Generic masculine detection}: Flags texts using the German generic masculine form.
    \item \textbf{Gender-neutral language detection}: Identifies inclusive writing styles such as gender colons or stars (e.g., \textit{Lehrer:innen}).
    \item \textbf{PMI adjectives}: Extracts the ten adjectives with the highest Pointwise Mutual Information (PMI) per actor, providing insights into recurring descriptive patterns.
    \item \textbf{Yearly aggregation}: Metrics are aggregated per year to enable longitudinal analysis of shifts in gendered representation and framing.
    \item \textbf{Yearly report generation}: All extracted metrics are compiled into a structured, human-readable report for each year.
\end{compactitem}

This approach, however, remains purely descriptive. Our work extends it substantially: we introduce new actor-level discrimination metrics and integrate a two-stage exclusion framework to move from diagnosis to corpus correction.

\section{The Extended Actor-Centred Pipeline}
\label{sec:extended}
We extend the actor-level pipeline introduced by \citet{urchs-etal-2024-detecting} and scaled in \citet{urchs2025taz2024fullanalysinggermannewspapers} to improve both analytical granularity and corpus curation. Unlike classical bias detection methods, which rely on document-level labels or aggregate statistics, our approach captures discrimination at the level of individual actors, making structural inequalities within texts visible.

Building on systemic functional linguistics~\cite{Halliday2004-ck} and critical discourse analysis~\cite{reisigl2017}, the pipeline incorporates metrics targeting key dimensions of discursive inequality:

\begin{compactitem}
    \item \textbf{Syntactic roles}: Distinguishing subject and object positions provides a proxy for agency. Actors in subject roles are framed as active agents, while object roles position them as passive. Tracking this distribution across gender groups highlights structural asymmetries in agency~\cite{Halliday2004-ck}.
    \item \textbf{Naming vs.\ pronoun reference}: Whether actors are referred to by name or reduced to pronouns affects their individuation and visibility. Persistent differences between genders can signal unequal treatment in how actors are foregrounded~\cite{larcher2015linguistische}.
    \item \textbf{Quotation style}: Direct quotations attribute voice and authority, while indirect quotations background speakers. Measuring the ratio of direct to indirect speech shows how discursive authority is distributed~\cite{larcher2015linguistische}.
    \item \textbf{Sentiment}: The evaluative framing of actors, captured via sentiment analysis, indicates whether certain groups are systematically associated with more negative language.
    \item \textbf{Pointwise Mutual Information (PMI)}: By extracting strongly associated adjectives, verbs, and nouns, we reveal the thematic and lexical contexts in which actors are embedded, surfacing stereotypical associations.
\end{compactitem}

These metrics go beyond frequency counts to capture framing, which classical fairness metrics (e.g., demographic parity) overlook. Actor-level analysis adds value over methods such as hate speech classification or keyword-based stereotype detection by revealing who is made visible, who is granted agency or voice, and how evaluations differ across gender groups. Insights that word- or document-level approaches cannot provide.

The pipeline outputs structured reports that combine these metrics with summary statistics, enabling both qualitative and quantitative inspection. It also supports a two-stage user-centred filtering mechanism: (1) flagging articles with strong internal asymmetries, and (2) rebalancing overall gender ratios. This ensures that the resulting corpus is not only analysed but also curated to reduce discriminatory patterns. The full pipeline code, documentation, and yearly reports are available at \url{https://github.com/Ognatai/corpus_balancing}

\section{Pipeline Application: Discrimination Analysis and Corpus Balancing}

We apply the extended actor-centred pipeline in two stages: first for diagnostic analysis, then for corrective balancing. Detection alone is insufficient: if left uncorrected, strong asymmetries risk skewing corpus statistics and reinforcing discriminatory patterns in downstream applications. Our pipeline, therefore, combines analysis with systematic filtering and balancing.

\subsection{Stage 1: Discrimination Analysis Across the Corpus}

In the first stage, the pipeline processes all articles and computes the actor-level metrics described in Section~\ref{sec:pipeline} and Section~\ref{sec:extended}. Results are aggregated per article and year to enable both fine-grained inspection and diachronic analysis. Yearly reports combine the full set of metrics in a structured, interpretable format, supporting both quantitative tracking of trends and qualitative exploration of framing practices (see Appendix~\ref{app:report} for an example). Actors are only tract per text, not in the whole corpus.

\subsection{Stage 2: Multi-Stage Filtering and Corpus Balancing}
\label{subsec:balancing}
The pipeline first produces a histogram showing, for all articles, the proportion of actors coded with she/her pronouns and the proportion of their mentions, ranging from 0\% (only he/him) to 100\% (only she/her). This initial view allows users to inspect the distribution of gender ratios before any intervention and to set thresholds for four asymmetry indicators introduced in Section~\ref{sec:pipeline}: \textit{sentiment gap}, \textit{subject/object ratio}, \textit{quote imbalance}, and \textit{naming imbalance} (named versus pronoun mentions).

Each indicator is computed per article for the two groups and compared as a \textit{ratio difference} with +1 Laplace smoothing to stabilise small counts. Concretely, for group $g \in \{\textit{she}, \textit{he}\}$ we define
\begin{equation}
    \textit{subject/object}(g) = \frac{\textit{subjects}_g + 1}{\textit{objects}_g + 1}
\end{equation}

\begin{equation}
    \textit{direct/indirect}(g) = \frac{\textit{direct}_g + 1}{\textit{indirect}_g + 1}
\end{equation}

\begin{equation}
    \textit{named/pronoun}(g) = \frac{\textit{named}_g + 1}{\textit{pronoun}_g + 1}
\end{equation}

and \textit{sentiment}(g) is the article level average polarity for mentions of group $g$. An article is flagged on a given indicator if the absolute difference between the two group-specific values exceeds a user-chosen threshold. Users also specify the minimum number of indicators that must be triggered simultaneously for an article to be excluded.

Thresholds are chosen with two principles in mind. First, \textit{sentiment} operates in a narrow numeric range around neutrality, so even moderate absolute differences are meaningful for evaluative framing. Second, the structural ratios \textit{subject/object}, \textit{direct/indirect}, and \textit{named/pronoun} exhibit higher natural variability across topics and genres, therefore stricter cut offs help avoid false positives from incidental fluctuations. Intuitively, a large ratio difference marks a sustained structural tilt, for example a pattern where one group appears predominantly as grammatical subjects relative to objects, is quoted directly rather than paraphrased, or is referred to by name rather than by pronoun, compared with the other group. This configuration enables flexible yet principled flagging, and Section~\ref{sec:balance} reports the concrete threshold values used in this study together with their empirical motivation.

After text-level exclusion, a second histogram is generated to show the updated distribution of gender ratios across articles. At this stage, the user decides on an equilibrium range for corpus-level balancing by specifying lower and upper bounds (e.g., how much more men can appear than women, and vice versa). 

Corpus-level balancing then iteratively excludes articles that contribute most to the remaining imbalance until actor- and mention-based ratios fall within the chosen range. A final histogram visualises the adjusted distribution and documents the effect of the balancing step.

Finally, all excluded article IDs are consolidated, and a revised balanced corpus is created. It is saved in the same format as the original dataset, but as a new version, ensuring compatibility while providing a fairer foundation for downstream use.

\section{Corpus-Balancing of \texttt{taz2024full}}
\label{sec:balance}

We use the \texttt{taz2024full} corpus~\cite{urchs2025taz2024fullanalysinggermannewspapers}, comprising over 1.8 million articles from the German left-leaning newspaper \textit{taz} (1980–2024). In the unfiltered corpus, we detect female- and male-coded actors in 1,834,018 articles. Actor frequency peaks in 2004 with 23,580 actors (7,523 female and 16,057 male). In early years, coverage is sparse and dominated by a small number of actors, but from 1988 onwards the corpus broadens significantly.  

\subsection{Imbalances Before Filtering}

\begin{figure}[ht]
    \centering
    \includegraphics[width=\linewidth]{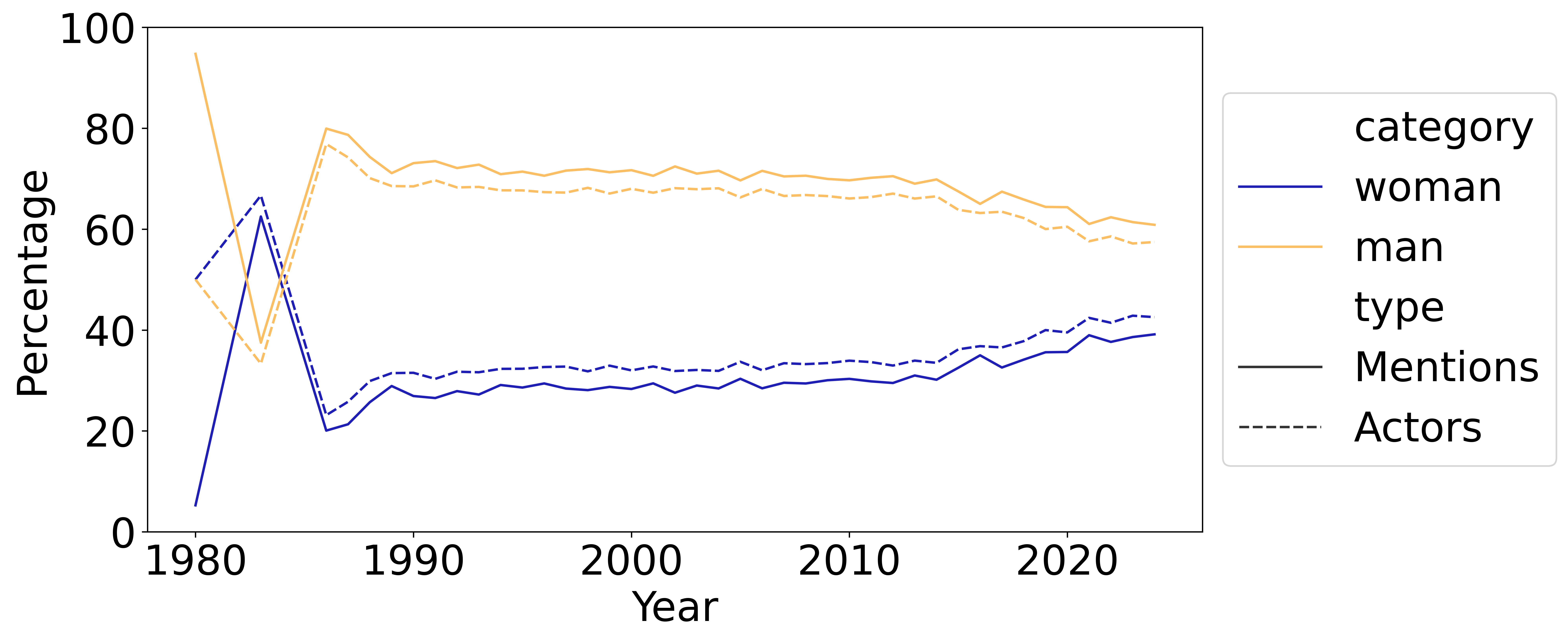}
    \caption{Percentage of male- and female-coded references over time \textit{before filtering}. Fluctuations in the early years reflect the small number of available articles.}
    \label{fig:combined_mentions_actors_before}
\end{figure}

\begin{figure}[ht]
    \centering
    \includegraphics[width=\linewidth]{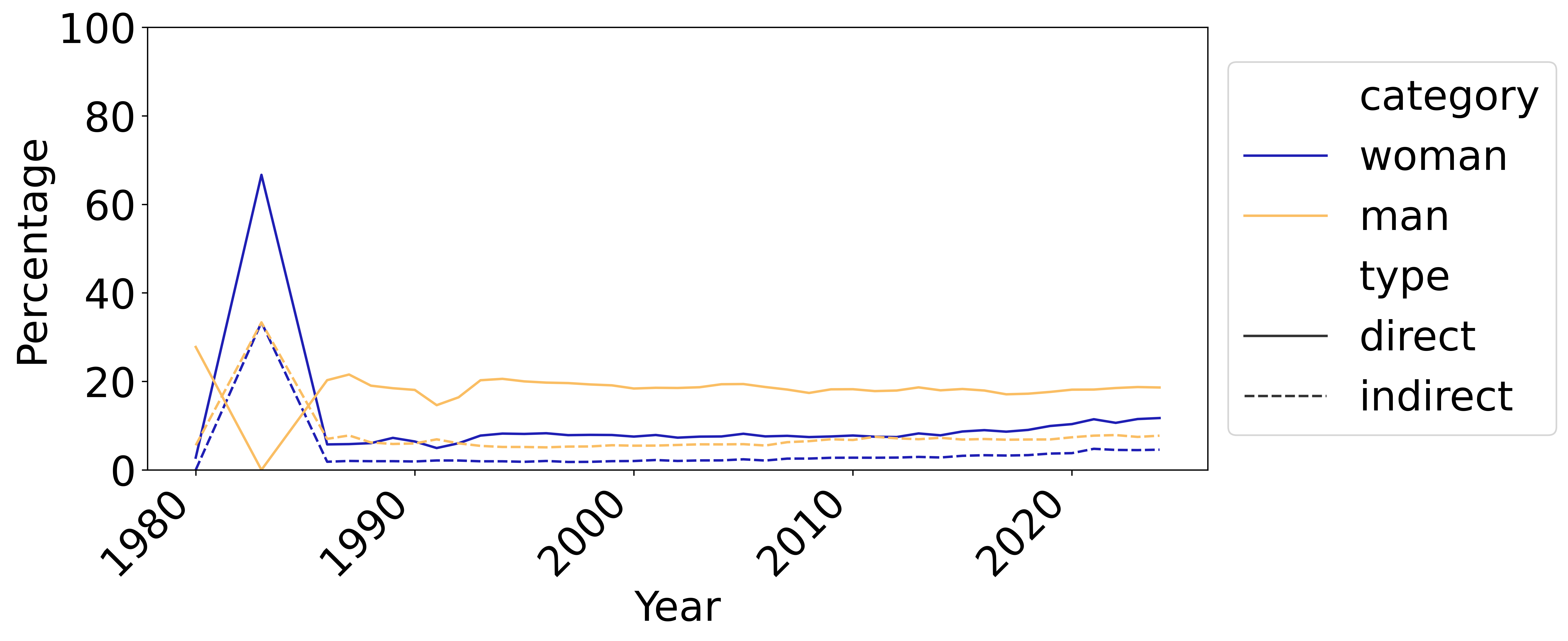}
    \caption{Distribution of quotation styles by gender \textit{before filtering}. Early-year fluctuations are attributable to low article counts.}
    \label{fig:direct_vs_indirect_before}
\end{figure}

\begin{figure}[ht]
    \centering
    \includegraphics[width=\linewidth]{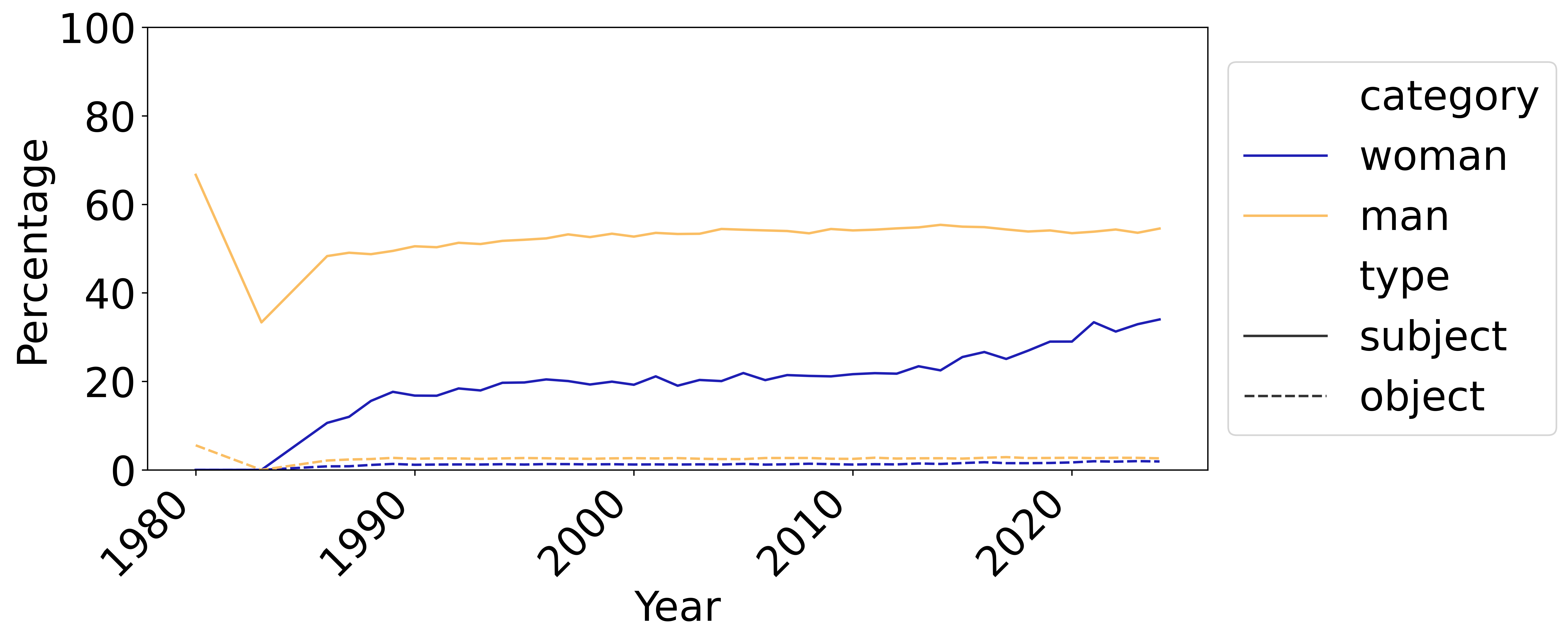}
    \caption{Distribution of syntactic roles by gender \textit{before filtering}. Early-year fluctuations are attributable to low article counts.}
    \label{fig:subject_vs_object_before}
\end{figure}

Across the unfiltered corpus, men dominate both actor counts and mention frequencies (Figure~\ref{fig:combined_mentions_actors_before}). These asymmetries are reflected not only in absolute representation but also in discursive positioning. Men appear more often in subject roles (cf. Figure~\ref{fig:subject_vs_object_before}) and as speakers in direct quotations (cf. Figure~\ref{fig:direct_vs_indirect_before}), while women are more frequently placed in object positions (cf. Figure~\ref{fig:subject_vs_object_before}) or paraphrased through indirect quotes (cf. Figure~\ref{fig:direct_vs_indirect_before}).  

\begin{figure}[ht]
    \centering
    \includegraphics[width=\linewidth]{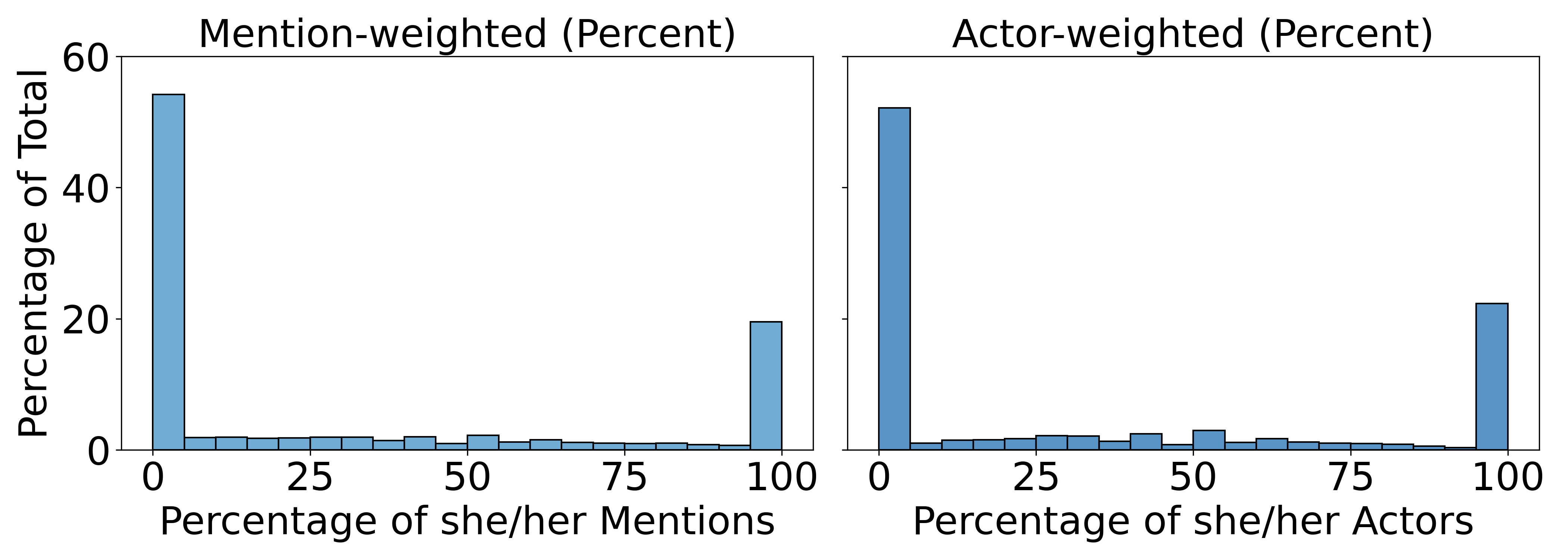}
    \caption{Distribution of gender ratios across articles \textit{before filtering}.}
    \label{fig:ratio_before}
\end{figure}

At the article level, gender representation is highly polarised: many texts reference either only male-coded or only female-coded actors (Figure~\ref{fig:ratio_before}). This shows that imbalance is not simply an aggregate effect but is embedded in the composition of individual articles.  

\subsection{Asymmetry Flags}

During the first text-level filtering step, we excluded 20 articles using four asymmetry flags:
\textit{sentiment gap}, \textit{quote imbalance}, \textit{subject/object ratio}, and \textit{naming imbalance}
(named vs.\ pronoun mentions). We decided to trigger the document exclusion if two or more flags were detected in a text to prevent over-exclusion.

Each flag compares \textit{ratio differences} between female- and male-coded actors with +1 Laplace smoothing to avoid division by zero, and fires when the absolute difference exceeds a preset threshold: sentiment gap $> 0.3$ (difference in average polarity), subject/object ratio difference $> 0.5$, direct/indirect quote ratio difference $> 0.5$, and named/pronoun mention ratio difference $> 0.5$. The cut-off values were chosen to capture asymmetries that go beyond natural stylistic variation and that are likely to affect how actors are framed in discourse. For \textit{sentiment}, a relatively low threshold of 0.3 was used, since polarity scores are generally close to neutral and even moderate differences can shift evaluative framing. For \textit{subject/object roles}, \textit{quoting}, and \textit{naming}, we required larger divergences of 0.5 in ratio space. These features are structurally more variable across texts, and a stricter cut-off ensures that only sustained imbalances are flagged.

\paragraph{Flag overlap.} Co-occurrence analysis shows that only twenty of the texts exhibit multiple asymmetries simultaneously. Most excluded texts (17,212) are flagged for a single imbalance, primarily subject/object distribution. 564 articles did not trigger any of the four asymmetry flags and were removed in the subsequent corpus-level balancing step to bring the overall actor and mention ratios into the target equilibrium.

\begin{figure}[ht]
    \centering
    \includegraphics[width=\linewidth]{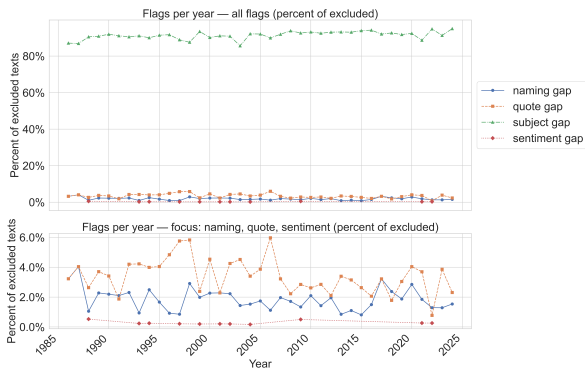}
    \caption{Proportion of excluded texts per year by flag type. Subject-role asymmetry dominates, while naming, quoting, and sentiment gaps occur less frequently.}
    \label{fig:flags_per_year}
\end{figure}

\paragraph{Flag frequencies.} Figure~\ref{fig:flags_per_year} shows the share of excluded texts per year by flag type. The \textit{subject gap} dominates throughout the corpus, consistently accounting for more than 80\% of flagged texts across all decades. This stability suggests that structural asymmetries in grammatical agency are a persistent feature of the newspaper’s coverage rather than a phenomenon tied to specific periods. The other three indicators occur more rarely, together contributing less than 10\% of exclusions. The \textit{quote gap} shows the most variation over time: it reaches values of up to 5-6\% of excluded texts in the 1990s and early 2000s, but remains lower and more stable after 2010. These spikes may reflect topic-specific reporting practices in those decades, such as an emphasis on political debates or international conflicts where male actors dominated as attributed speakers, while female actors were more often paraphrased. The \textit{naming gap} occurs at low levels (1-2\%) without a clear temporal trend, while the \textit{sentiment gap} is negligible throughout, with only a slight increase visible after 2010. Overall, no systematic long-term trends are observable beyond the persistent dominance of subject-role asymmetries and the temporary spikes in quoting imbalances around the turn of the millennium.

\paragraph{Qualitative examples.} To illustrate these asymmetries, we include examples from the excluded set:  
\begin{compactitem}
    \item \textbf{Subject-object gap:} In a football report on Eintracht Frankfurt, all named actors are male and consistently appear as grammatical subjects: \textit{“Horst Ehrmantraut [...] gelang es, mit geringen finanziellen Mitteln den Aufstieg zu realisieren”}, \textit{“Rolf Heller [...] regiert heute auf dem Präsidentenstuhl”}, \textit{“Weber [...] hat nach langem Pokern einen neuen Vierjahresvertrag unterschrieben”}. \\
    Female-coded actors are entirely absent from the text, reinforcing an imbalance where men hold agency in the discourse while women do not appear as subjects at all.
    
    \item \textbf{Quote imbalance:} In a political portrait of Peter-Michael Diestel, male actors are repeatedly given direct speech: 
    \textit{“Alle, alle, waren da und wollten mich haben [...]”}, \textit{“Ich bin strunzbieder. Ich bin ein Konservativer. Ich stehe zum CDU-Programm.”}, \textit{“Vor Schröder hätte er Schiß gehabt.”} \\
    Female-coded actors, by contrast, are only mentioned collectively (e.g. \textit{“Eppelmann, Heitmann und andere [...]”}) and paraphrased without direct quotations.
  
   \item \textbf{Naming gap:} In a film review, the female protagonist is repeatedly introduced by name: \textit{“Deniz, die 21-jährige Heldin in Arslans Film, geht an diesem nicht enden wollenden Sommertag Rohmer-Filme synchronisieren.}, \textit{"Wie Deniz an diesem Tag ihre Wäsche zur Mutter bringt, die Schwester trifft, ihren Freund verlässt [...]”} \\
   Male figures in the same text, such as her boyfriend or the director Thomas Arslan, are mentioned once and then largely referred to with pronouns. 
   
    \item \textbf{Sentiment gap:} In a letter to the editor, female actors are explicitly evaluated in negative terms, for example: \textit{“[...] wie ist es möglich, dass die Autorin ohne Kommentar oder Richtigstellung wahrheitswidrig schreiben kann [...]”} and dismissively mocked: \textit{“Sie sind herzlich eingeladen, für ihre hehren Werte mit einer Menschenkette an der Front [...] zu demonstrieren.”} \\
    By contrast, male commentators in the same article (e.g. Hartmut Rosa, Peter Bethke, Gerhard Harms) are described neutrally or respectfully.
\end{compactitem}

\subsection{Global Equilibrium}

In a second step, we applied corpus-level balancing, excluding an additional 17,816 articles to bring the overall actor and mention ratios into the target range $[0.75, 1.25]$. This interval means that one gender may occur up to 25\% more frequently than the other: for example, a ratio of 1.25 indicates that female actors or mentions outnumber male ones by 25\%, while a ratio of 0.75 indicates the reverse. Both directions are treated symmetrically, ensuring that neither male nor female dominance persists beyond this margin. The choice of range enforces approximate parity without creating an artificial 1:1 distribution, while retaining authentic temporal dynamics.

\begin{figure}[ht]
    \centering
    \includegraphics[width=\linewidth]{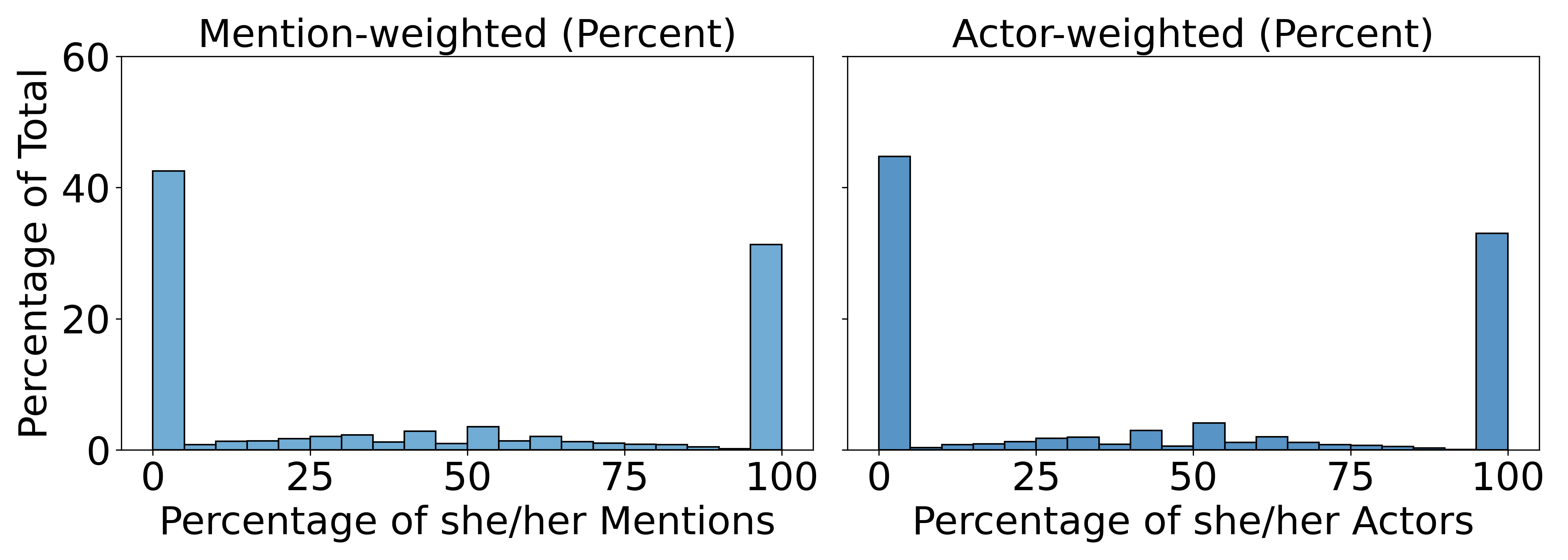}
    \caption{Distribution of gender ratios across articles \textit{after corpus-level balancing}. The x-axis shows the percentage of she/her references (mentions on the left, actors on the right). The y-axis shows the proportion of texts. Peaks at 0\% and 100\% are strongly reduced after balancing, indicating that one-gender-only articles were downsampled.}
    \label{fig:ratio_after_balancing}
\end{figure}

Figure~\ref{fig:ratio_after_balancing} shows the resulting distribution of gender ratios across all articles. Compared to the unfiltered corpus (cf. Figure~\ref{fig:ratio_before}), the distribution is more centred and less polarised. Articles with exclusively male-coded or exclusively female-coded actors, which previously created sharp spikes at 0\% and 100\%, have been reduced. Instead, more texts now fall into the mid-range, where both genders are present. This demonstrates that global balancing successfully decreased the extreme ends of the distribution while preserving variation in the middle range.

\paragraph{Excluded texts.} The equilibrium step removed texts that were maximally polarised in their gender representation. Across all 17,796 excluded articles, we detect \textbf{35,995 male-coded actors} and \textbf{190,192 male-coded mentions}, but no female-coded actors or mentions. In other words, every article excluded by this step contained only men. Such one-sided texts were frequent enough to skew the corpus-level balance if left untouched, producing systematic over-representation of male-coded actors.

\begin{figure}[ht]
    \centering
    \includegraphics[width=\linewidth]{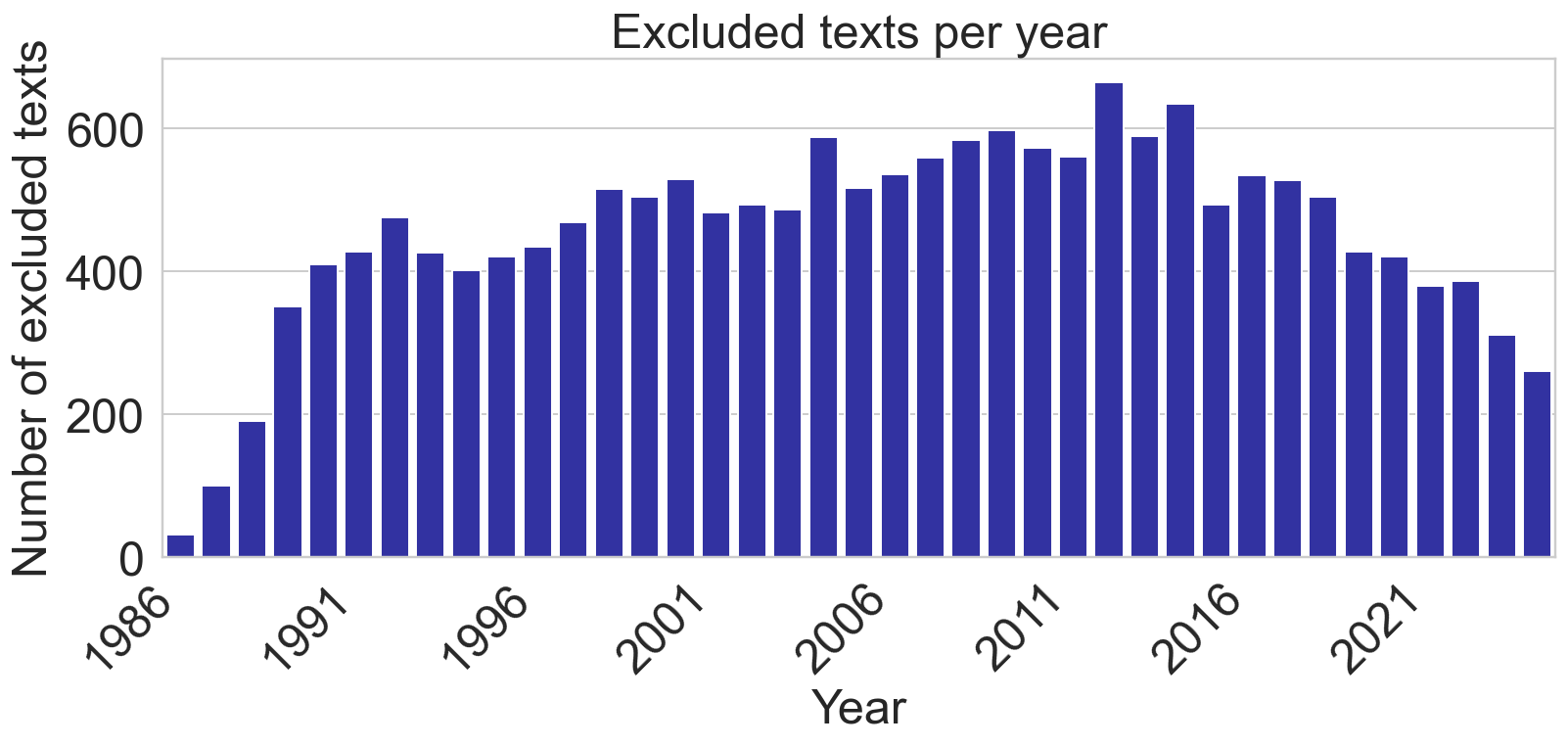}
    \caption{Number of excluded texts per year after all exclusion steps (text-level filtering and corpus-level balancing). Exclusions scale with article production and are distributed across the corpus history.}
    \label{fig:excluded_per_year}
\end{figure}

\paragraph{Temporal distribution.} Exclusions occur across the entire corpus history (Figure~\ref{fig:excluded_per_year}), but their frequency closely tracks overall article production. In the late 1980s and early 1990s, very few articles are excluded, reflecting the limited size of the corpus at that stage. From the mid-1990s onwards the number of exclusions rises steadily, stabilising at around 400–500 per year. Between 2005 and 2015 exclusions remain consistently high, often exceeding 500 texts annually, with a clear peak around 2015 (over 650). After 2018, the numbers decline again, falling below 300 in the most recent years. This pattern indicates that exclusions are not confined to the early, sparse period, but accompany phases of high article production and decline in line with overall corpus dynamics. Furthermore, we could detect no temporal trend in discrimination.

\subsection{Results After Filtering and Balancing}

\begin{figure}[ht]
    \centering
    \includegraphics[width=\linewidth]{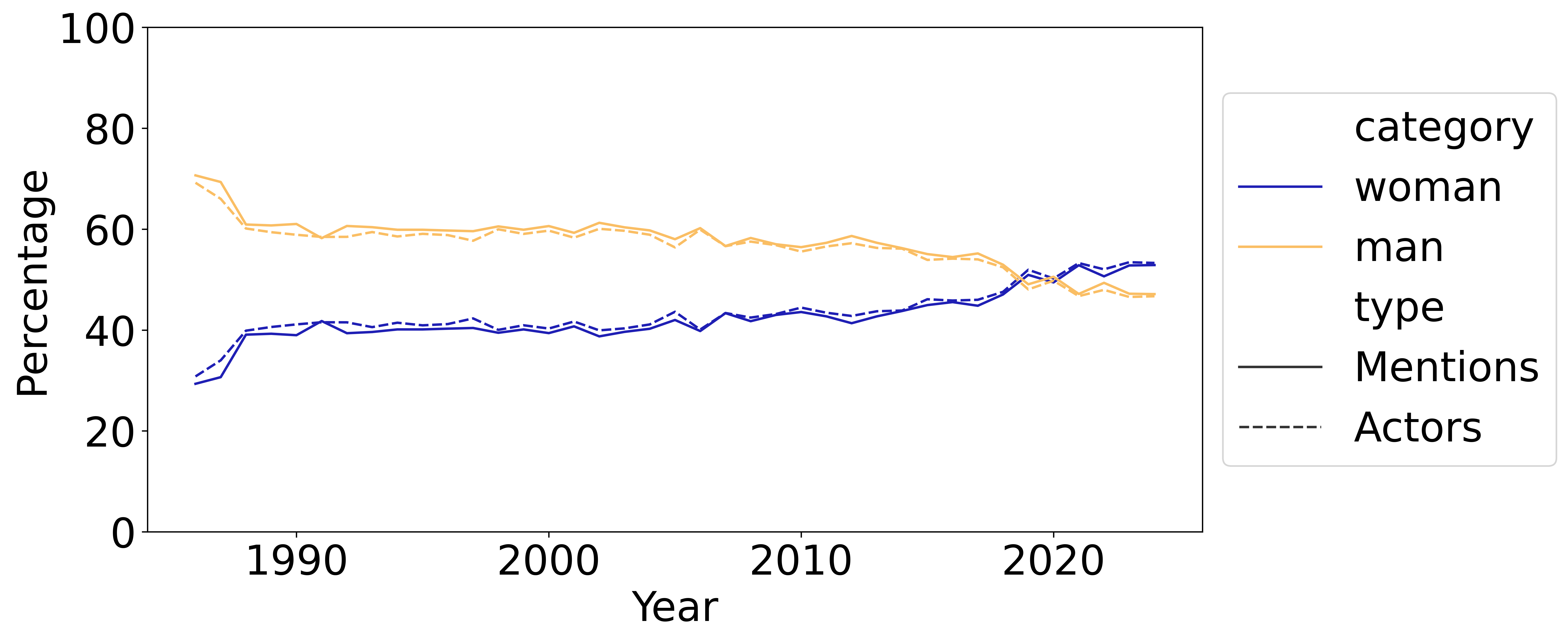}
    \caption{Percentage of male- and female-coded references over time \textit{after all filtering and balancing steps}. The trajectories of mentions and actors converge towards parity while retaining the natural crossing point around 2018.}
    \label{fig:combined_mentions_actors_after}
\end{figure}

The final corpus exhibits near parity across mentions and actor counts (Figure~\ref{fig:combined_mentions_actors_after}). Importantly, structural dynamics such as the crossing point around 2018 remain intact, indicating that balancing improves representation without erasing genuine historical patterns. Compared to the unfiltered corpus (cf.\ Figure~\ref{fig:combined_mentions_actors_before}), the trajectories of male- and female-coded actors now run in parallel, showing that referential balance has been restored across time.

\begin{figure}[ht]
    \centering
    \includegraphics[width=\linewidth]{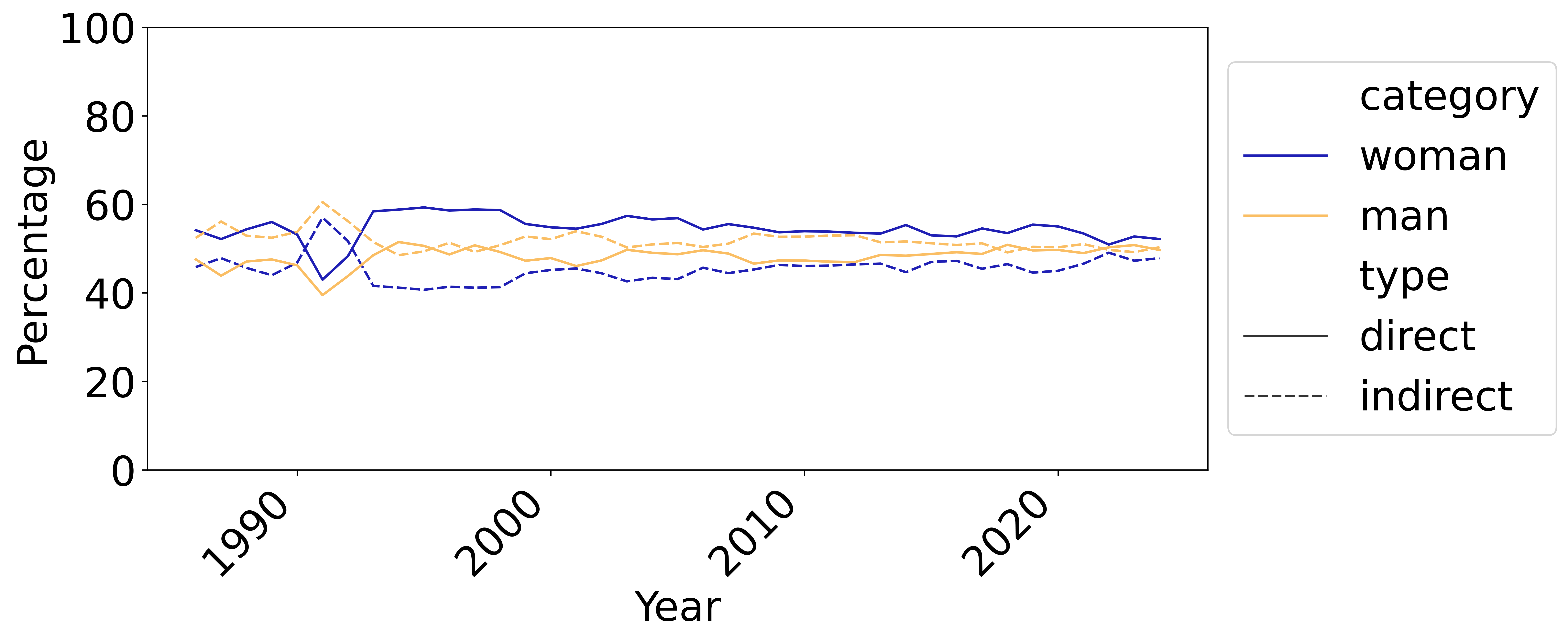}
    \caption{Proportion of direct and indirect quotations by gender \textit{after full exclusion}. The gap is reduced compared to the unfiltered corpus, with women more frequently quoted directly.}
    \label{fig:direct_vs_indirect_after}
\end{figure}

Figure~\ref{fig:direct_vs_indirect_after} shows the distribution of quotation styles. After balancing, women appear more often in direct speech than before, reducing the quote imbalance observed in Figure~\ref{fig:direct_vs_indirect_before}. Men still receive slightly more indirect quotations, but the gap is narrower, suggesting that women’s discursive agency is more strongly represented in the final corpus.

\begin{figure}[ht]
    \centering
    \includegraphics[width=\linewidth]{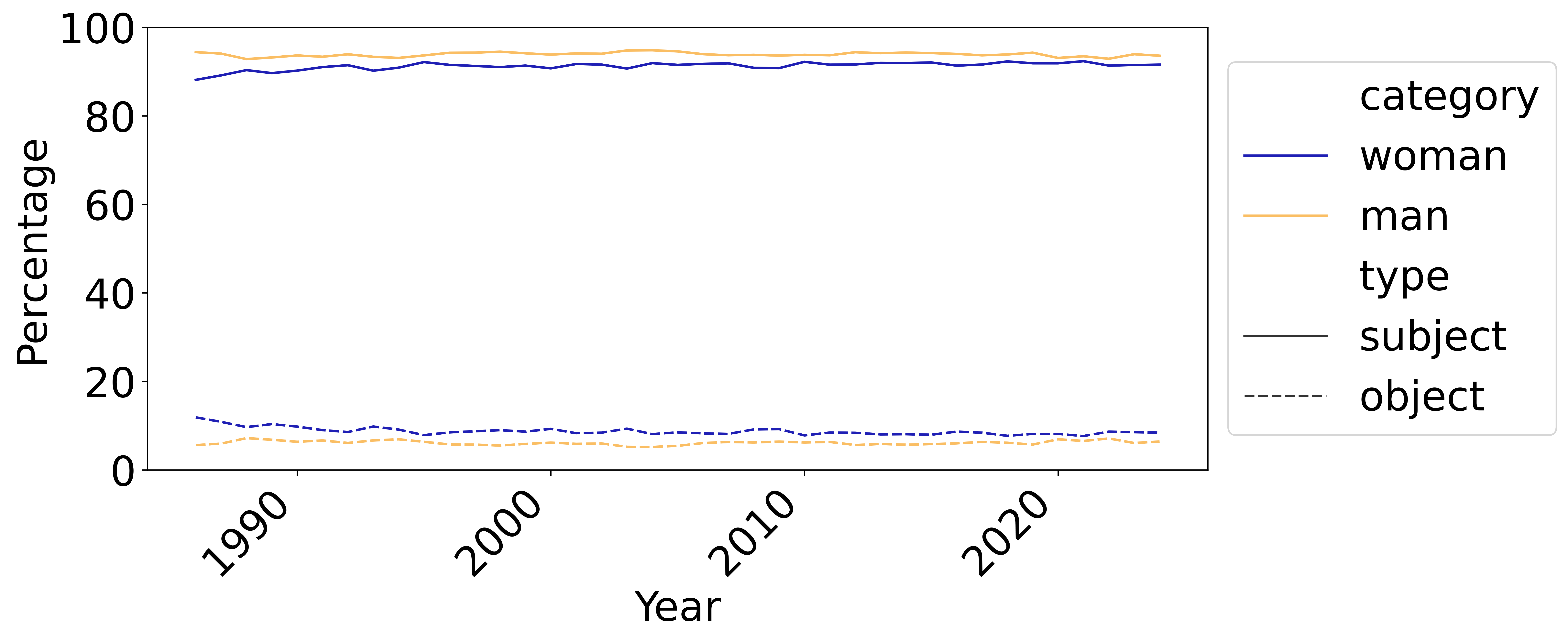}
    \caption{Distribution of syntactic roles \textit{after full exclusion}. The subject–object gap between men and women is markedly reduced compared to the unfiltered corpus.}
    \label{fig:subject_vs_object_after}
\end{figure}

Finally, Figure~\ref{fig:subject_vs_object_after} illustrates the distribution of syntactic roles. Men still occur more frequently in subject positions, but the difference is markedly reduced compared to the unfiltered corpus (cf. Figure~\ref{fig:subject_vs_object_before}). The gap decreases from around 30 percentage points to roughly 5, showing that grammatical agency is now distributed more evenly across genders. This represents one of the strongest structural improvements achieved by the balancing process.

Taken together, these figures demonstrate that the corpus is not only numerically balanced but also structurally improved. Referential parity is achieved, women are quoted more often in their own words and syntactic agency is redistributed more evenly. The balancing process thus mitigates multiple dimensions of gender inequality while preserving historically meaningful variation.

\section{Conclusion and Future Work}

We presented an extended actor-level pipeline for detecting and mitigating gender discrimination in large-scale text corpora. Beyond prior work, we introduced metrics for syntactic roles, quotation, and sentiment, structured reports for interpretability, and a two-stage filtering process for building more balanced corpora. 

Applied to the \texttt{taz2024full} corpus, our approach shows that gender imbalances in representation and framing are both measurable and correctable. The resulting corpus is more balanced across multiple linguistic dimensions and provides a stronger foundation for corpus-based analysis and fairer NLP practices. 

Yet some asymmetries, particularly in implicit discourse structures, persist. Future work should address these through context-aware models, targeted debiasing strategies, and intersectional extensions that include race, age, and class. Expanding actor categories beyond the gender binary will further support inclusive analysis. More broadly, we argue that discourse-aware methods should become part of corpus construction workflows, as understanding how groups are framed is essential for designing fairer NLP systems.

\section*{Use of AI}
The authors are not native English speakers; therefore, ChatGPT and Grammarly were used to assist with writing English in this work. ChatGPT was also used to assist with coding.

\section*{Limitations}
While our approach enables corpus-level balancing based on measurable framing asymmetries, it has limitations. The exclusion strategy reduces corpus size and may remove valuable content alongside biased texts. It also relies on surface-level linguistic signals and cannot capture subtler biases such as irony, omission, or topic choice. Furthermore, the method enforces a binary gender classification, excluding non-binary identities, and it applies only to texts with identifiable actors and gender cues, leaving some material outside the analysis.

\section*{Ethical Considerations}

Our work is grounded in the belief that fairness in NLP requires not only technical interventions but also critical reflection on the social impact of language technologies. By analysing how gendered actors are represented and framed in text, we make structural inequalities visible and address them at the level of data design. Yet fairness cannot be reduced to numerical balance: filtering texts entails normative choices about which content is deemed discriminatory, with risks of over-correction and loss of context. Our reliance on binary gender resolution further excludes non-binary and gender-nonconforming individuals, reinforcing the very simplifications we seek to critique. We consider this a significant ethical limitation and aim to extend our methods to more inclusive representations. Finally, while we mitigate discrimination in training data, responsibility also lies in model architectures, deployment contexts, and the socio-technical systems in which NLP tools operate.

\section*{Acknowledgments}

Matthias Aßenmacher is funded with funds from the Deutsche Forschungsgemeinschaft (DFG, German Research Foundation) as part of BERD@NFDI - grant number 460037581. Responsibility for the contents of this publication lies with the authors.

\bibliography{corpus_balancing}

\clearpage

\section*{Appendix}

\appendix

\section{Corpus Report 2023}
\label{app:report}

\noindent\begin{minipage}{\textwidth}
    \centering
    \lstset{
        numbers=none,
        language=,
        basicstyle=\small\ttfamily,
        frame=none
    }
    \begin{lstlisting}

Report for the year 2023
===========================================================================

AGGREGATED TOTALS (all texts)
Total Texts:                        10019
Texts with Actors:                  10019
Uses Gender Neutral Language (Docs):  107
Generic Masculine Usage (Docs):      8081

Metric                                  she/her      he/him     overall
---------------------------------------------------------------------------
Pronoun Distribution:                      6892        9194       16086
Mentions by Pronoun:                      35595       56044       91639
Named Mentions:                           22544       36047       58591
Pronoun Mentions:                         13051       19997       33048
Subject Roles:                            18625       30303       48928
Object Roles:                              1119        1540        2659
Direct Quotes:                             6501       10588       17089
Indirect Quotes:                           2529        4215        6744
Feminine-coded Words:                      4251        6066       10317
Masculine-coded Words:                     2870        4764        7634
Sentiment:                                -0.01       -0.01       -0.01
Named Mentions (% of all mentions):        38.5        61.5            
Pronoun Mentions (% of all mentions):        39.5        60.5            
Subject Roles (% of known roles):          38.1        61.9            
Object Roles (% of known roles):           42.1        57.9            
Direct Quotes (% of quotes):               38.0        62.0            
Indirect Quotes (% of quotes):             37.5        62.5            

STATISTICS (per text)
---------------------------------------------------------------------------
Metric                                             Mean    Median   Std Dev
---------------------------------------------------------------------------
Pronouns (Resolved) (She/Her)                      0.69      1.00      0.83
Mentions (By Pronoun) (She/Her)                    3.55      2.00      7.22
Feminine Coded Words (By Pronoun) (She/Her)        0.42      0.00      1.18
Masculine Coded Words (By Pronoun) (She/Her)       0.29      0.00      0.77
Named Mentions (Sum Over Actors) (She/Her)         2.25      1.00      5.65
Pronoun Mentions (Sum Over Actors) (She/Her)       1.30      1.00      2.29
Subject Roles (She/Her)                            1.86      0.00      3.78
Object Roles (She/Her)                             0.11      0.00      0.43
Direct Quotes (She/Her)                            0.65      0.00      1.46
Indirect Quotes (She/Her)                          0.25      0.00      0.72
Pronouns (Resolved) (He/Him)                       0.92      1.00      0.91
Mentions (By Pronoun) (He/Him)                     5.59      3.00      9.77
Feminine Coded Words (By Pronoun) (He/Him)         0.61      0.00      1.36
Masculine Coded Words (By Pronoun) (He/Him)        0.48      0.00      1.05
Named Mentions (Sum Over Actors) (He/Him)          3.60      1.00      7.75
Pronoun Mentions (Sum Over Actors) (He/Him)        2.00      1.00      3.02
Subject Roles (He/Him)                             3.02      2.00      5.01
Object Roles (He/Him)                              0.15      0.00      0.53
Direct Quotes (He/Him)                             1.06      0.00      1.96
Indirect Quotes (He/Him)                           0.42      0.00      0.97
Mean Sentiment (All)                              -0.02      0.00      0.10
Total Actors                                       1.61      1.00      1.04
Total Mentions                                     9.15      5.00     12.10
Total Feminine Coded Words                         1.03      0.00      1.83
Total Masculine Coded Words                        0.76      0.00      1.28
Uses Gender-Neutral Language                       0.01      0.00      0.10
Generic Masculine                                  0.81      1.00      0.40       
    \end{lstlisting}
\end{minipage}

\clearpage

\noindent\begin{minipage}{\textwidth}
    \centering
    \lstset{
        numbers=none,
        language=,
        basicstyle=\footnotesize\ttfamily, 
        frame=none
    }
    \begin{lstlisting}
        TOP PMI ADJECTIVES
---------------------------------------------------------------------------------------
Most frequent adjectives associated with each pronoun group.

Rank ALL                           she/her                      he/him                        
---------------------------------------------------------------------------------------
1    letzten (414.00)              letzten (154.00)             letzten (269.00)              
2    russischen (272.00)           junge (130.00)               russischen (195.00)           
3    deutschen (260.00)            berliner (101.00)            deutschen (171.00)            
4    berliner (231.00)             deutschen (97.00)            politische (142.00)           
5    junge (212.00)                deutsche (97.00)             ukrainische (137.00)          
6    nächsten (212.00)             russischen (81.00)           politischen (135.00)          
7    politische (212.00)           nächsten (80.00)             berliner (134.00)             
8    deutsche (208.00)             politischen (80.00)          nächsten (133.00)             
9    politischen (205.00)          politische (74.00)           ukrainischen(117.00)         
10   ukrainische (178.00)          jungen (71.00)               russische (113.00)            


TOP PMI NOUNS
---------------------------------------------------------------------------------------
Most frequent nouns associated with each pronoun group.

Rank ALL                           she/her                       he/him                        
---------------------------------------------------------------------------------------
1    menschen (588.00)             menschen (311.00)             menschen (315.00)             
2    frau (353.00)                 frau (234.00)                 präsident (289.00)            
3    präsident (328.00)            frauen (163.00)               mann (210.00)                 
4    leben (312.00)                leben (140.00)                partei (185.00)               
5    mann (280.00)                 mutter (128.00)               leben (182.00)                
6    partei (268.00)               kinder (109.00)               land (164.00)                 
7    land (238.00)                 tochter (107.00)              frau (147.00)                 
8    frauen (210.00)               geschichte (101.00)           sohn (135.00)                 
9    stadt (209.00)                mann (100.00)                 stadt (135.00)                
10   regierung (208.00)            anfang (100.00)               mittwoch (126.00)             


TOP PMI VERBS
---------------------------------------------------------------------------------------
Most frequent verbs associated with each pronoun group.

Rank ALL                           she/her                       he/him                        
---------------------------------------------------------------------------------------
1    erzählt (671.00)              erzählt (331.00)              erzählt (368.00)              
2    steht (495.00)                steht (199.00)                steht (324.00)                
3    sieht (449.00)                erklärt (180.00)              sieht (315.00)                
4    erklärt (428.00)              lassen (167.00)               erklärt (269.00)              
5    lassen (359.00)               sieht (163.00)                erklärte (243.00)             
6    erklärte (346.00)             sehen (147.00)                spricht (228.00)              
7    spricht (341.00)              zeigt (139.00)                lassen (205.00)               
8    zeigt (302.00)                spricht (139.00)              sprach (199.00)               
9    weiß (289.00)                 lebt (127.00)                 zeigt (190.00)                
10   hält (286.00)                 sagen (125.00)                weiß (188.00)   
    \end{lstlisting}
\end{minipage}

\end{document}